\documentclass[journal]{IEEEtran}
\usepackage{comment}
\usepackage{amsmath}
\usepackage{mathtools}
\usepackage{multirow}
\usepackage{amssymb}
\usepackage{booktabs}
\usepackage{algorithmic}
\usepackage[ruled]{algorithm}

\usepackage{soul}
\usepackage{xcolor}
\usepackage{subfig}
\usepackage{url}
\usepackage{cite}

\begin{document}
%
\title{A Deep Learning Driven Algorithmic Pipeline for Autonomous Navigation in Row-Based Crops}
%
%
%

\author{
Simone Cerrato,
Vittorio Mazzia,
Francesco Salvetti,
Mauro Martini,
Simone Angarano,
Alessandro Navone,
Marcello Chiaberge

\thanks{All the authors are with the Department of Electronics and Telecommunications (DET), Politecnico di Torino, Torino, TO 10129, email: name.surname@polito.it}

\thanks{This work has been developed with the contribution of the Politecnico di Torino Interdepartmental Centre for Service Robotics (PIC4SeR).}
}

\maketitle

\begin{abstract}
Expensive sensors and inefficient algorithmic pipelines significantly affect the overall cost of autonomous machines. However, affordable robotic solutions are essential to practical usage, and their financial impact constitutes a fundamental requirement to employ service robotics in most fields of application. Among all, researchers in the precision agriculture domain strive to devise robust and cost-effective autonomous platforms in order to provide genuinely large-scale competitive solutions.
In this article, we present a complete algorithmic pipeline for row-based crops autonomous navigation, specifically designed to cope with low-range sensors and seasonal variations. Firstly, we build on a robust data-driven methodology to generate a viable path for the autonomous machine, covering the full extension of the crop with only the occupancy grid map information of the field. Moreover, our solution leverages on latest advancement of deep learning optimization techniques and synthetic generation of data to provide an affordable solution that efficiently tackles the well-known Global Navigation Satellite System unreliability and degradation due to vegetation growing inside rows. Extensive experimentation and simulations against computer-generated environments and real-world crops demonstrated the robustness and intrinsic generalizability to different factors of variations of our methodology that opens the possibility of highly affordable and fully autonomous machines.
\end{abstract}


%
\IEEEpeerreviewmaketitle

\section{Introduction}
\label{sec:intro}
Agriculture 3.0 and 4.0 have gradually introduced autonomous machines and interconnected sensors into several agricultural processes \cite{mazzia2021deepway, salvetti2022waypoint}, trying to introduce robust and cost-effective novel solutions into the overall production chain. For instance, precision agriculture has progressively innovated tools for automatic harvesting \cite{roldan2018robots}, vegetative assessment \cite{zhang2020assessment}, crops yield estimation \cite{feng2020yield}, smart and sustainable pesticide spraying robotic system \cite{deshmukh2020design} and many others \cite{khaliq2019refining, radoglou2020compilation}. Indeed, the pervasiveness of precision agriculture techniques has such a huge impact on certain fields of application that the adoption of them has become increasingly essential to achieve high product quality standards \cite{comba2015vineyard}. Moreover, the introduction of robots in agriculture will increasingly have a stronger impact on economic, political, social, cultural, security, \cite{sparrow2021robots}, and will be the only tool to satisfy the future food demand of our society \cite{duckett2018agricultural}.

Nevertheless, research on autonomous machines still requires further developments and improvements to meet the necessary industrial conditions of robustness and effectiveness. Global path planning \cite{mazzia2021deepway, salvetti2022waypoint},  mapping \cite{garg2021semantics}, localization \cite{saeedi2018navigating} and decision-making \cite{mota2020commonsense} are only some of the required tools that each year undergo heavy research from the scientific community to achieve the necessary requirements for full automation. Among all requirements, a low financial impact constitutes a fundamental goal in order to provide genuinely large-scale competitive solutions \cite{levoir2020high}. Indeed, expensive sensors and demanding computational algorithms significantly impact the actual usefulness of robotics platforms, essentially preventing their large-scale adoption and introduction into the agricultural world. 

\begin{figure}[!t]
\centering
\includegraphics[width=1.0\columnwidth]{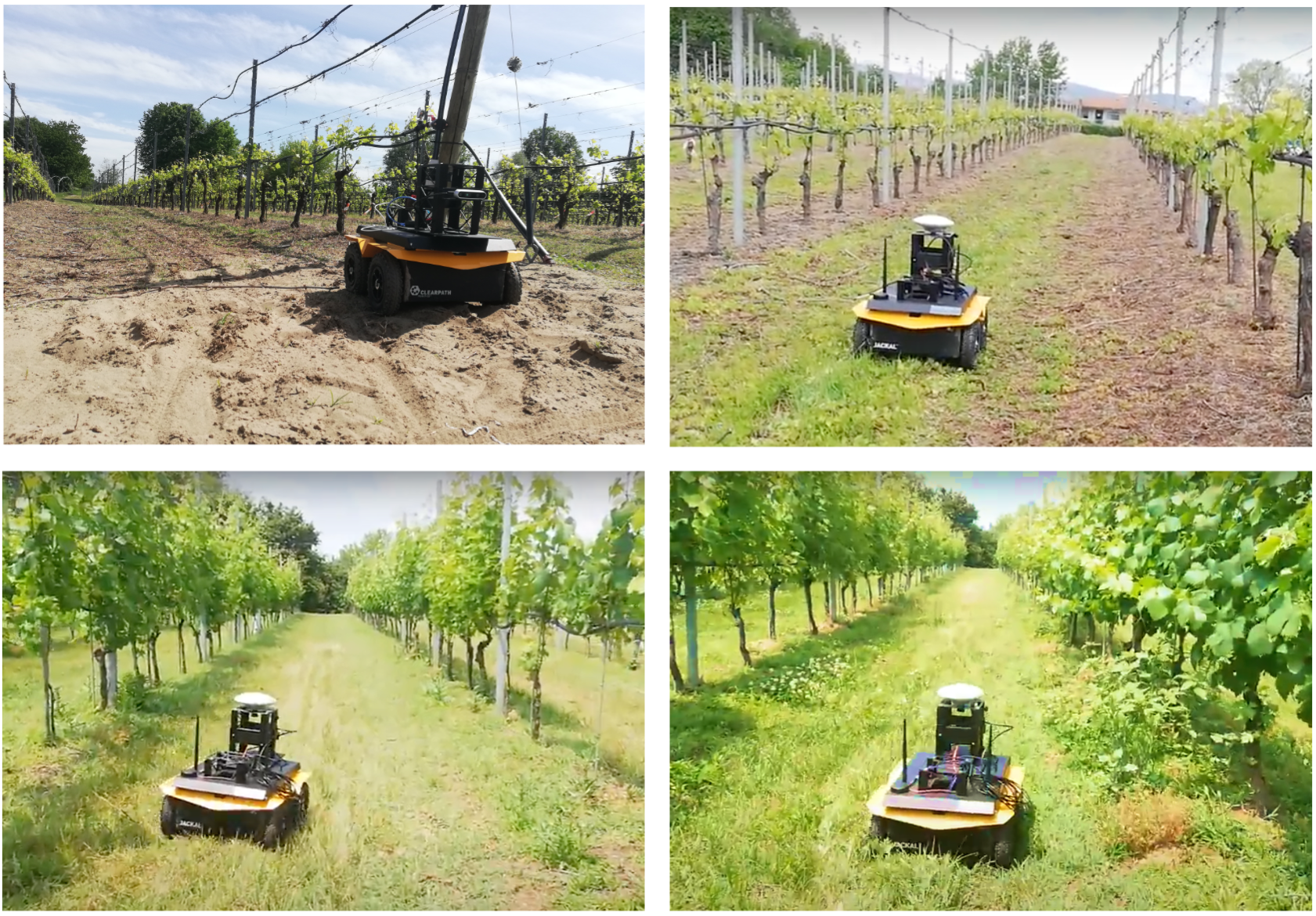}
\caption{Field tests with the Jackal platform in different seasonal periods of the same crop. Lush vegetation and thick canopies greatly reduce the GPS accuracy, affecting its reliability and  consequently the overall navigation pipeline. Nevertheless, our proposed segmentation-based algorithm exploits semantic segmentation properties to provide a proportional controller that drives the robotic platform along the whole row.}
\label{fig:seasonal_variation}
\end{figure}

Recently, deep learning methodologies \cite{hinton2015nature} revolutionized the entire computer perception field, endowing machines with an unprecedented representation of the surrounding environment \cite{grigorescu2020survey}. Moreover, the intrinsic robustness of representation learning techniques to different factor of variations opens the possibility to achieve noteworthy perception capabilities with low-cost and low-range sensors, greatly relieving the overall cost of the target machine \cite{boschi2020cost, martini2022position}. Finally, the latest advancement in deep learning optimization techniques, \cite{jacob2018quantization}, have progressively reduced latency, inference cost, memory, and storage footprint of its algorithms. That effectively scales down computational requirements enabling low-power computational devices boosted by hardware accelerators such as visual processing units (VPUs), tensor processing units (TPUs), and embedded GP-GPUs, \cite{mazzia2020real}.

Building on the latest deep learning researches for computer perception and exclusively making use of low-range sensors, we present a complete algorithmic pipeline for autonomous navigation in row-based crops. The proposed robust solution is explicitly designed to adapt to seasonal variation, as shown in Fig. \ref{fig:seasonal_variation}, ensuring complete coverage of the field in the different situations. Moreover, the low financial impact of our methodology greatly reduces maintenance and production costs and enables a large-scale adoption of fully autonomous machines for precision agriculture.

Firstly, we build on a robust data-driven methodology to generate a viable path for the autonomous machine, covering the full  extension  of  the  crop  with  only  the  occupancy  grid map  information  of  the  field. Successively, depending on the vegetation growth status of the crop, we adopt a purely Global Navigation Satellite System (GNSS) local path planning or a vision-based algorithm that exclusively makes use of a low-cost RGB-D camera to navigate inside the inter-row space. Indeed, meteorological conditions and especially lush vegetation and thick canopies, significantly affect GNSS reliability, degrading its precision and consequently the overall navigation pipeline, \cite{kabir2016performance, marden2014gps}. Conversely, our vision-based system exploits semantic information of the environment to navigate between rows and depth information to refine the underline control smoothness, disentangling from the necessity of a precise localization. Moreover, we make exclusively usage of synthetic data and domain randomization, \cite{tobin2017domain}, to enable affordable supervised learning and simultaneously bridging the domain gap between simulation and reality. Such technique allows us to easily construct and train a deep learning model able to efficiently generalize on different row-based crops. 

The overall proposed methodology guarantees to autonomously navigate throughout row-based crops without expensive sensors and with every seasonal variation.
All the algorithmic pipeline has been developed ROS-compatible in order to make easier the communication among different software modules and to be easily deployed on our developing
platform, the Jackal Unmanned Ground Vehicle (UGV) by
Clearpath Robotics\footnote{https://clearpathrobotics.com/}. Extensive experimentation and simulations against computer-generated environments and diverse real-world crops demonstrated the robustness and intrinsic generalizability of our solution. All of our code\footnote{https://github.com} and data\footnote{https://zenodo.com}
are open source and publicly available. 
\vspace{-7pt}
\subsection{Related Work}
Over the past years, autonomous systems designed to navigate in row-crops fields make largely usage of high-precision GNSS
receivers  in combination with laser-based sensors \cite{ly2015fully, moorehead2012automating}. Nevertheless, the canopies
on the sides of the row reduce the GNSS accuracy, affecting
its reliability and forcing the adoption of more expensive sensors \cite{kabir2016performance}. Indeed, more robust solutions fuse multiple sensor information (GNSS,  inertial navigation systems, wheel encoders) to obtain a better estimation of the mobile platform location in presence of thick canopies and adverse meteorological conditions \cite{overview_auto_nav_UGV, winterhalter2021localization}. However, high adoption of high-range sensors leads to higher system and maintenance costs, preventing a large-scale adoption of self-driving agricultural machinery. 

On the other hand, visual odometry (VO), \cite{zaman2019cost, nevliudov2021development}, and computer vision based solutions, \cite{auto_nav_for_wolfberry, semantic_seg_synthetic_images}, have been proposed as more affordable approaches. Nonetheless, VO
systems show poor performance on long distances due to the
accumulating error and struggle with highly similar patterns \cite{kim2018low} and unpredictable lighting conditions \cite{anthony2017uav}. Moreover, purely vision-based solutions cannot deal with seasonal variations and generalizability to different crops is difficult to achieve.

\begin{figure*}[t]
\centering
\includegraphics[width=\textwidth]{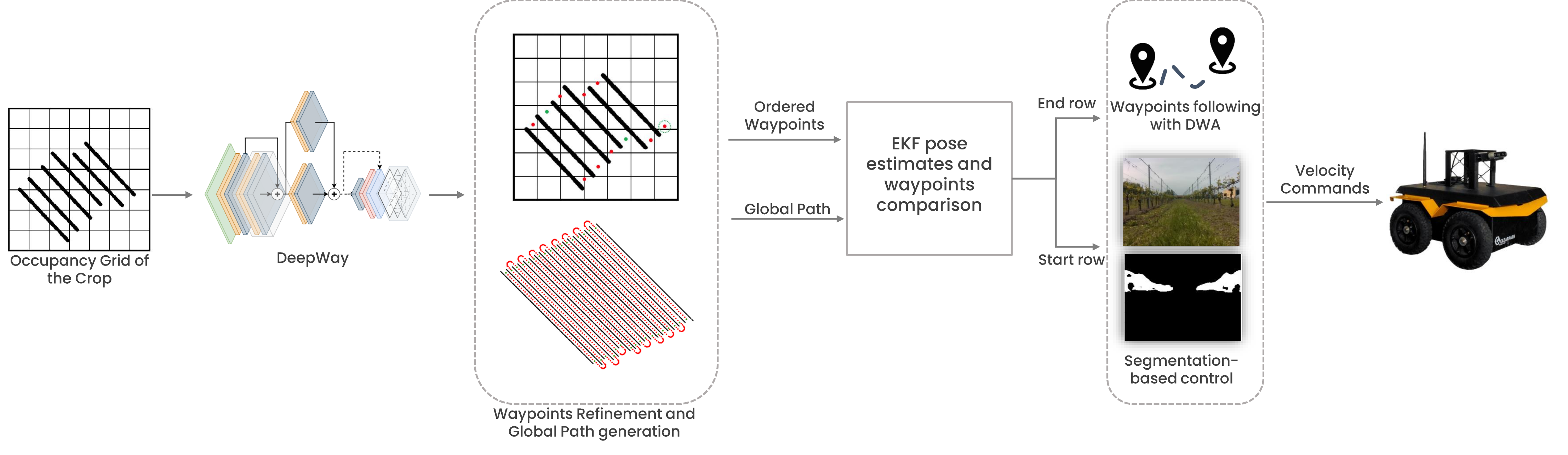}
\caption{A representation of the overall pipeline. From left to rigth, an occupancy grid of the crop is provided as input to DeepWay neural network, that estimates the waypoints at start/end of vineyards row. Then, a custom algorithm is responsible to order the generated waypoints and compute a global path maintaining a safe distance from crops. Finally, a local path planning policy chooses the right navigation algorithm according to UGV's position with respect to waypoint in order to drive the mobile platform along the whole path.}
\label{fig:tot_system}
\end{figure*}

In this state-of-the-art landscape, our team started in 2019 a research project with the precise aim to develop a complete working pipeline to autonomously navigate in vineyard rows without relying on multiple expensive sensors. As already introduced, computer vision and deep learning based algorithms, \cite{hinton2015nature}, have demonstrated particular robustness in solving problems with noisy signals. Moreover, optimization and edge AI techniques have progressively made inference computational affordable, opening the usage of deep learning methodologies to diverse practical applications  \cite{kamilaris2018deep}. Consequently, in \cite{aghi2020autonomous, aghi2021deep} were addressed navigation in vineyard rows with deep learning based algorithm and the exclusive usage of an RGB-D camera. Furthermore, starting with \cite{zoto2019automatic} it has been addressed the  global path generation automation problem, which is commonly neglected by the research community. However, a suitable path generator is crucial for obtaining a complete autonomous navigation performance and its absence prevents the control of the platform on the field. For that reason, in \cite{mazzia2021deepway}, moving from the clustering solution proposed in \cite{zoto2019automatic}
to detect the rows of the vineyards, DeepWay, a robust data-driven approach for global path generation, was presented. Besides being more robust and easier to adopt, DeepWay is a highly general approach that can be extended to every kind of row-based crops.
\vspace{-7pt}
\subsection{Novelties}
Based upon the aforementioned previous work, the main contributions of the presented approach herein are as follows:
\begin{itemize}
    \item Introduction of a highly affordable complete algorithmic pipeline for autonomous navigation in row-based crops. The methodology makes use of only low-cost, low-range sensors to drive a platform for the full extension of a crop in the different seasonal periods.
    \item Building on \cite{aghi2021deep}, we extend the local navigation between vineyard rows to a general row-based crop. We propose a domain randomization based training procedure to easily obtain a segmentation network with only synthetic data. 
    \item We introduce a novel global path planning procedure to connect two successive rows, avoiding the usage of too general algorithms that could easily introduce jerky and inefficient paths. 
    \item Improved usage of semantic information of the crop to navigate between rows in case of lush vegetation and thick canopies.
\end{itemize}

Compared to other existing autonomous navigation algorithms our solution makes use of low-cost sensors as: a cheap GNSS receiver with Real-Time Kinematic (RTK) corrections, an RGB-D camera, an Inertial Measurement Unit (IMU) and encoders. Our approach aims to fill the gaps between usage of low-cost sensors and robustness of autonomous navigation in the precision agriculture context exploiting the collaboration between Artificial Intelligence and standard navigation algorithms. On the other hand, existing solution exploits high cost and very accurate sensors, as: 3D LiDAR, expensive RTK-GNSS receiver to achieve a reliable and robust complete autonomous solution.

The remainder of this paper is organized as follows. Section \ref{system_overview} describes the overall system framework. The detailed explanation of the proposed full pipeline is introduced in Section \ref{full_pipeline}. Section \ref{algorithm_testing} presents the pipeline evaluation either against computer-generated environments or real-world crops. Finally, Section \ref{conclusions} draws conclusions and suggests future work.

\section{System Overview}

\label{system_overview}
The proposed work is intended for presenting a complete autonomous navigation system for general row-based crops. The designed autonomous system is organized in several software modules that should collaborate with each other in order to obtain effective and reliable driverless navigation throughout the whole field. A visualization of the complete pipeline is shown in Fig. \ref{fig:tot_system}.

Firstly, the system takes as input a georeferenced occupancy grid map of the considered crop to compute a global path made of geographic coordinates; in particular, it exploits the DeepWay network, \cite{mazzia2021deepway},  to estimate the start/end waypoints of each row, then a custom global path planner, \cite{cerrato2021adaptive}, computes the desired path maintaining a safe distance from crops. Secondly, according to the season period and the amount of crop vegetation, it is possible to choose the kind of navigation to perform: only GNSS-based or GNSS and AI-assisted. 

In case a good view of sky is available both outside and inside the row space, the system exploits only Real Time Kinematic (RTK) corrections, GNSS signals, and inertial data to autonomously guide the mobile platform throughout the crops. On the other hand, when lush vegetation is present, the proposed navigation system makes use of RTK corrections, GNSS signals, and inertial data to perform the row switch, since outside the row space a good sky view is available, while along the rows it exploits the camera, the deep neural network, and the segmentation-based control to overcome GNSS signals unreliability. 
In both scenarios, the system strongly relies on estimated global positions of the UGV in order to follow the provided global path. In our approach, the localization problem is tackled fusing the positioning information coming from an RTK enabled GNSS receiver and the inertial data provided by an IMU. All the data is loosely fused exploiting the well-known Extended Kalman Filter (EKF), that uses an omnidirectional model for prediction and outputs position estimations in the form of: $x, y, yaw$, since the navigation happens in 2-dimension. $x $ and $y$ are global spatial information represented in the East-North-Up (ENU) reference frame, while $yaw$ is the absolute orientation with respect to magnetic north, corrected with the actual magnetic declination.
Once the localization filter is set up, in case a clear view of the sky is available, the navigation algorithm uses the estimated UGV positions, the global path and a local planner based on the Dynamic Window Approach (DWA) to autonomously guide the mobile platform; otherwise, it makes use of both GNSS-based and AI-assisted navigation, that exploit GNSS signals and semantic information of crops to safely navigate in the whole field without colliding with the crops.
In the latter case, the navigation type selection occurs comparing the estimated UGV global positions and an ordered waypoints list computed by the DeepWay neural network and successively refined. Finally, it is important to underline that DWA navigation scheme is only one possible solution. It is adopted in the presented pipeline for its simplicity, flexibility and online collision avoidance capability. However, further experimentation with even simpler algorithms has been performed but not presented for the sake of conciseness. For instance, the Pure Pursuit controller \cite{coulter1992implementation} demonstrated very promising results between and outside rows, bringing possible advantages in presence of more packed rows or  larger vehicles.

\section{Methodology}
\label{full_pipeline}
\begin{figure}[!t]
\centering
\includegraphics[width=0.7\columnwidth]{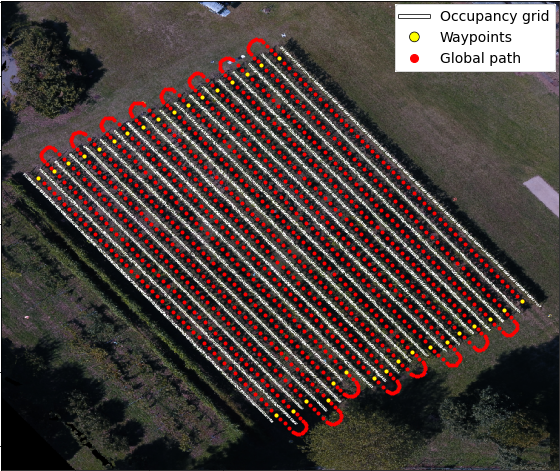}
\caption{Aerial view of a row crop field, together with the occupancy grid (white), the estimated waypoints (yellow), and the global path (red). The meters/pixel resolution is 0.1 m/px, the end-row distance margin $d_{er} = 20$ px, that results in a 2 meters real-world margin.}
\label{fig:map_example}
\end{figure}

In this section, each block of the navigation system is presented, detailing all the aspects of the path planning and navigation processes. Our methodology requires as input a georeferenced occupancy grid of the target field $\textbf{\textit{X}}_{occ} \in  \mathbb{R}^{H\times W}$, obtained by segmenting an aerial view of the environment. The system is developed and tested on satellite imagery, but the very same methodology can be applied to images obtained by drones flying over the target field.  
The global path is computed by the first two blocks of the pipeline and is represented as an ordered set of points $\textbf{\emph{P}} = \{(x,y)| x,y \in \mathbb{R}\}$ to be followed by the UGV in order to reach full coverage of the target field. Fig. \ref{fig:map_example} shows an example of occupancy grid with the predicted global path in red superimposed on the aerial image of the field. Once the global path $\textbf{\emph{P}}$ has been generated, the system exploits a local path planning policy to autonomously navigate in the considered crop choosing the proper control algorithm according to seasonal period and the presence of thick canopies on crops. Indeed, full autonomous navigation happens exploiting the GNSS information and the semantic information obtained employing the segmentation network.

\vspace{-7pt}
\subsection{Waypoints Estimation}
The first block aims at predicting the list of $l$ waypoints ${\textbf{\emph{W}} \in \mathbb{N}^{l\times 2}}$ in the occupancy grid reference frame that represent the begin and end of each row of the target field. Since classical clustering methods fail with real-world conditions such as rows of different length, holes and outliers, we adopt the DeepWay framework \cite{mazzia2021deepway}, which frames the waypoints prediction as a regression problem. DeepWay is a fully convolutional neural network that takes as input the occupancy grid  $\textbf{\textit{X}}_{occ}$ of dimension $H \times W $ and outputs a map $\hat{\textbf{\textit{Y}}}$ of dimension $U_H \times U_W \times 3$:
\begin{equation}
    \hat{\textbf{\textit{Y}}} = f_{DeepWay}(\textbf{\emph{X}}_{occ})
    \label{eq:deepway}
\end{equation}
The first channel of $\hat{\textbf{\textit{Y}}}$ is a confidence map that outputs for each cell $u$ the probability $P(u)$ that a waypoint falls inside the cell itself. The output map dimensions are obtained subsampling the input space of a factor $k$:

\begin{equation}
    U_H = H/k \qquad U_W = W/k
\end{equation}

\begin{figure}[!h]
\centering
\includegraphics[width=0.95\columnwidth]{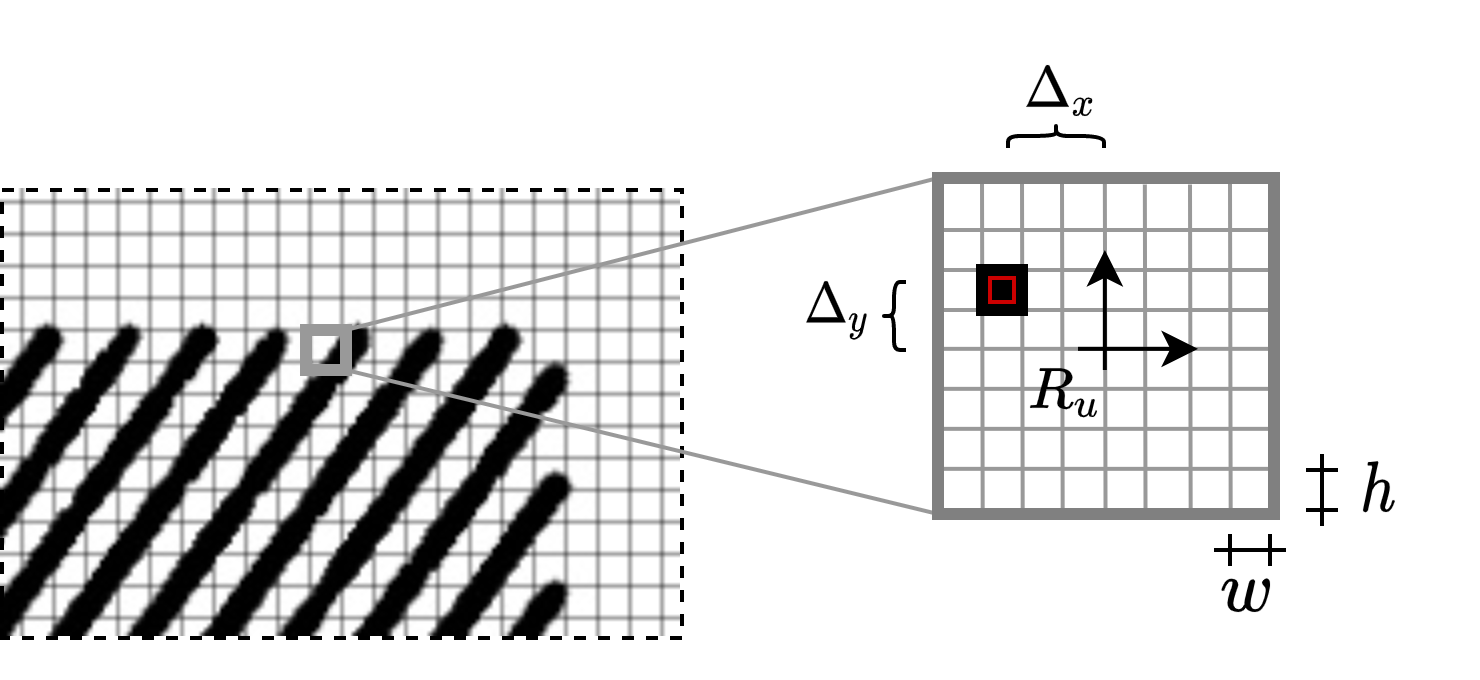}
\caption{Example of an output map $\hat{\textbf{\textit{Y}}}$ of DeepWay \cite{mazzia2021deepway}. Since $k=8$, each cell $u$ represents a square area of $8 \times 8$ pixels of the original occupancy grid. The local reference system $R_u$ is at the centre of the cell and the compensation factors $\Delta x$ and $\Delta y$ are normalized to the $[-1,1]$ range, as a fraction of the semi-cell length.}
\label{fig:grid_example}
\end{figure}

Thus, each cell $u$ represents a square region of $k \times k$ pixels of the original occupancy grid. The other two channels of the output map $\hat{\textbf{\textit{Y}}}$ predict for each cell $u$ two compensation factors $\Delta x$ and $\Delta y$ used to localize the waypoint inside the $u$ cell, as shown in Fig. \ref{fig:grid_example}. Those factors are normalized to $[-1,1]$ range so that they represent a distance from the center of the cell. Thus, a factor of 1 means a positive deviation on the corresponding axis of half the length of the cell. Eventually, the final location of the predicted waypoints in the input space can be recovered as:

\begin{equation}
    \hat{\mathbf{y}}_{O}=k\left(\hat{\mathbf{y}}_{U} + \frac{\mathbf{\Delta + 1}}{2}\right)
    \label{eq:wp_coor}
\end{equation}

where $\hat{\mathbf{y}}_{O}$ and $\hat{\mathbf{y}}_{U}$ are the vectors of $(x,y)$ coordinates of a generic waypoint in the input and output reference systems, respectively; $\mathbf{\Delta}$ is the vector of the $(\Delta x,\Delta y)$ normalized compensation factors.

The prediction confidences stored in the first channel of the output map $\hat{\textbf{\textit{Y}}}$ are compared to a confidence threshold $c_{thr}$, and all the positions with $P(u) > c_{thr}$ are selected and projected in the input space as in Eq. \ref{eq:wp_coor}. Furthermore, a suppression mechanism is adopted as in standard object detection algorithms in order to avoid multiple predictions of the same waypoint: all the points falling within a distance threshold $d_ {thr}$ from each other are replaced with the one with maximum confidence $P(u)$. The final waypoints are stored in the list ${\textbf{\emph{W}}}$.

The network is characterized by a stack of $N$ Residual Reduction modules, that are based on 2D convolutions with Mish activation \cite{misra2019mish} and implement both channel and spatial attention \cite{woo2018cbam}. Each module halves the spatial dimension with a Reduction block based on 2D convolutions with strides of two. After $N$ modules, a 2D Transpose Convolution increases the spatial dimension by a factor of two. A last 2D convolution projects a concatenation of the features of the last two modules to the 3-dimensional space of the output. Since the first and last convolutional layers also have strides of two, the output tensor spatial dimension is reduced by a factor $k = (N+1)^2$ with respect to the input. The final layer uses sigmoid activation for the first channel that encodes the waypoint probability and tanh activation for the other two channels that encode the normalized compensation factors. All code related to DeepWay is open-source and can be found online\footnote{https://github.com/fsalv/DeepWay}.

\subsection{Global Path Planning}
\label{sec:global_path_planning}
The output list of waypoints ${\textbf{\emph{W}}}$ should be ordered in order to plan a global path that reaches a full coverage of the field. We adopt the same post-processing as in \cite{mazzia2021deepway} with the following steps:

\begin{enumerate}
    \item the row crops orientation is estimated from the occupancy grid $\textbf{\textit{X}}_{occ}$ with the progressive probabilistic Hough transform \cite{matas2000robust}.
    \item the waypoints are clustered using the density-based algorithm DBSCAN\cite{ester1996density} that creates a variable number of clusters depending on the space density of the waypoints.
    \item the points in each cluster are ordered projecting them along the normal to the direction estimated in step 1).
    \item the clusters are merged with a heuristic approach based on their position and size until two main groups representing the two sides of the field are reached.
    \item the final ordered list of waypoints $\textbf{\emph{W}}_{ord} \in \mathbb{N}^{l\times 2}$ is obtained selecting the points from the two main clusters following an A-B-B-A scheme.
\end{enumerate}

The global path is generated from the ordered list $\textbf{\emph{W}}_{ord}$ in two steps. The intra-row paths are obtained with \cite{cerrato2021adaptive}, which exploits a gradient-based planner between the starting and the ending waypoints of each row. On the other hand, the inter-row paths are generated with a circular pattern in order to keep a safe margin from the end of the rows to avoid collision during the turns. Considering an end-row waypoint $\textbf{\textit{p}}_i$ and the successive point that starts the following row $\textbf{\textit{p}}_{i+1}$, the waypoints are firstly moved along the row direction to get an end-row margin $d_{er}$:

\begin{equation}
\begin{split}
    \textbf{\textit{p}}_i^{shifted} &= \textbf{\textit{p}}_i + d_{er} \begin{bmatrix} \text{cos}\alpha\\ \text{sin}\alpha\end{bmatrix}\\
    \textbf{\textit{p}}_{i+1}^{shifted} &= \textbf{\textit{p}}_{i+1} + (d_{er} + \Delta \textbf{\textit{d}})\begin{bmatrix} \text{cos}\alpha\\ \text{sin}\alpha\end{bmatrix} 
\end{split}
\end{equation}

where $\alpha$ is the angle estimated during the post-processing steps and $\Delta \textbf{\textit{d}}$ is the distance between the two points along the row direction:

\begin{equation}
    \Delta \textbf{\textit{d}} = (\textbf{\textit{p}}_i - \textbf{\textit{p}}_{i+1}) \cdot \begin{bmatrix} \text{cos}\alpha\\ \text{sin}\alpha\end{bmatrix} 
\end{equation}

The end-row margin $d_{er}$ can be selected depending on the meters/pixel resolution of the occupancy grid in order to have a target margin in meters in the real environment. A circular interpolation is adopted to connect the shifted points by linearly interpolating the angles considering the mean point as the center of the circumference.
The whole sequence of points obtained by the intra-row and inter-rows planning creates the global path $\textbf{\emph{P}} = \{(x,y)| x,y \in \mathbb{R}\}$, defined in the reference system of the occupancy grid $\textbf{\textit{X}}_{occ}$. If the field map is georeferenced, it is possible to convert the global path into a list of geographic coordinates that can be directly used in the local planning phase to control the UGV motion. In Fig. \ref{fig:map_example} an aerial view of a field is shown, together with the predicted waypoints in yellow and the global path in red.

\subsection{Segmentation Network}
\begin{figure}[!b]
\centering
\includegraphics[scale=0.63]{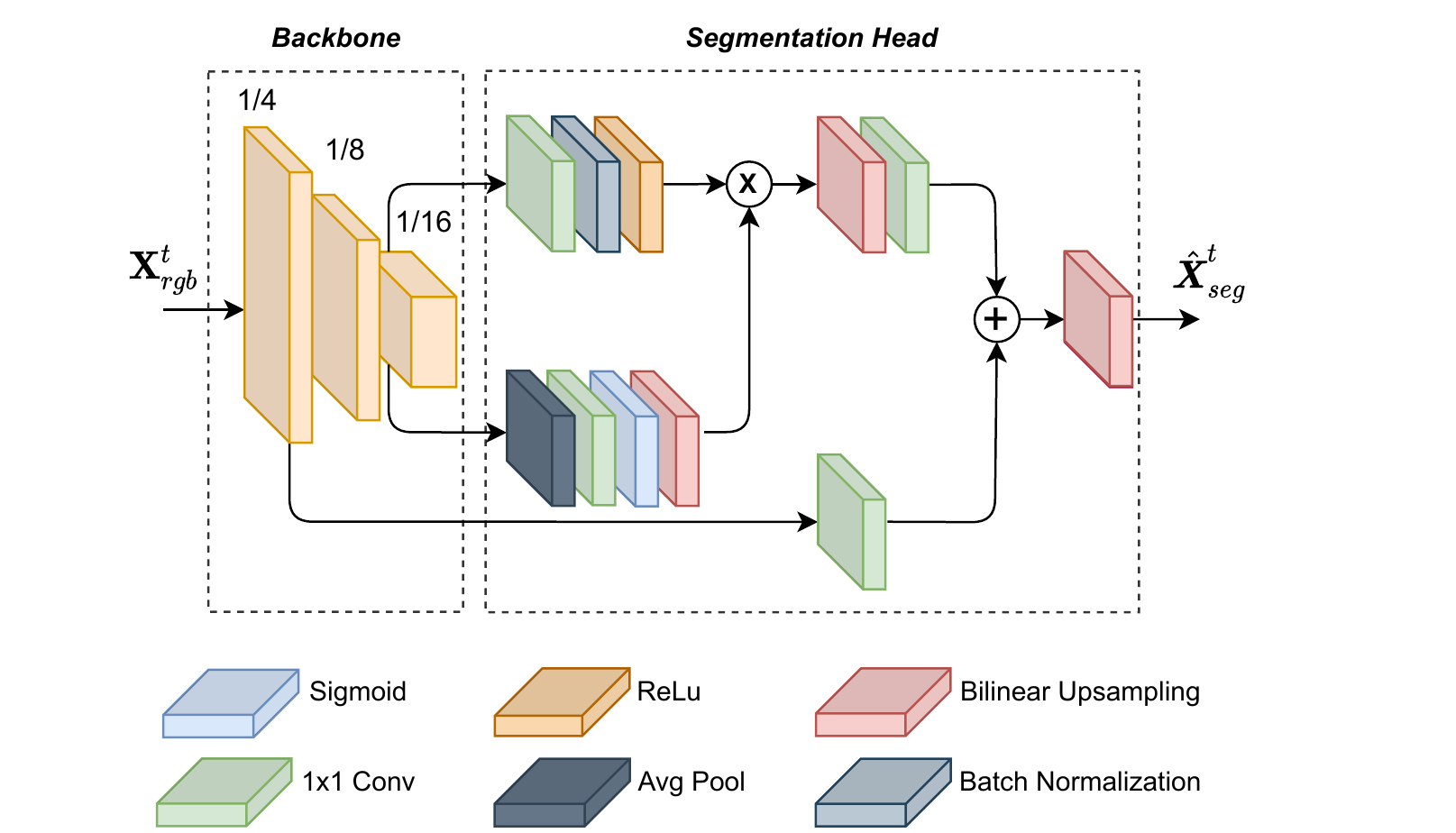}
\caption{Graphical representation of the architecture of the segmentation network. Features at different resolution feed the segmentation head that combines them producing the output binary map, $\hat{\textbf{\textit{X}}}_{seg}$.}
\label{fig:segnet_head}
\end{figure}
The overall segmentation network acts as a function $H_{seg}$, parameterized by $\Theta$, that at each temporal instant $t$ takes as input the RGB frame from the onboard camera $\textbf{\textit{X}}_{rgb} \in  \mathbb{R}^{h\times w\times c}$ and produces a binary map, $\hat{\textbf{\textit{X}}}_{seg} \in  \mathbb{R}^{h\times w}$ with $h$, $w$ and $c$ as height, width and channels, respectively. The output positive class segments the crops and the foliage in the camera view. Ideally, it should be equally split on the sides of the frame for a perfectly centered path. Successively, the semantic information of the row, $\hat{\textbf{\textit{X}}}_{seg}$, is used in conjunction with its corresponding depth map to control all movements of the platform inside the crops rows.

Among all recent real-time semantic segmentation models, we carefully select an architecture that guarantees high accuracy levels by also containing hardware costs, optimization simplicity, and computational load. Indeed, the segmentation-based control does not considerably benefit from fine grained predictions and elaborated encoder-decoder networks, \cite{hu2020real}, or two-pathway backbones, \cite{yu2020bisenet}, does not bring any considerable improvement. Therefore, we adopt a very lightweight backbone,  MobileNetV3 \cite{howard2019searching}, followed by a reduced
version of the Atrous Spatial Pyramid Pooling module, \cite{chen2017rethinking}, to capture richer contextual information with minimal computational impact. Indeed, the output of the last layer of the backbone can not be used directly to predict the segmentation mask due to the lack of spatial details.

The overall architecture is depicted in Fig. \ref{fig:segnet_head}. The backbone, with repeated spatial reductions, extracts contextual information and two of its branches at different resolution feed the segmentation head. One layer applies atrous convolution
to the 1/16 resolution to extract denser features, and the
other one is used to add a skip connection from the 1/4
resolution to work with more detailed information. Finally, in order to maintain real-time performance even without hardware accelerators, we employ a 224x224 low-resolution input. So, we rescale the global
average pooling layer setting the kernel size to 12×12 with
strides (4,5). Additionally, to have equal input and output
dimensions, we add a final bilinear upsampling with a factor of 8 at the end of the segmentation head.
\vspace{-5pt}
\subsection{Segmentation-based control}\label{seg_based_section}
The segmentation masks $\hat{\textbf{\textit{X}}}_{seg} \in  \mathbb{R}^{h\times w}$ provided by the deep neural network are post-processed and fed into a custom control algorithm in order to generate consistent velocity commands to drive the UGV inside the inter-row space and maintain as much as possible the inter-row centrality. As in \cite{aghi2021deep}, we compute a sum of $S$ segmentation maps along with an intersection with depth information provided by an RGB-D camera in order to obtain a more stable control. First, we pick $S$ consecutive segmentation maps at times $\{t-S,...,t\}$ and we fuse them 

\begin{equation}
    \hat{\textbf{\textit{X}}}_{cumSeg}^{t} = \sum_{n=0}^{S} \hat{\textbf{\textit{X}}}_{seg}^{t-n}
    \label{sum_seg_maps}
\end{equation}

then, we join the depth information $\textbf{\textit{X}}_{depth}^{t} \in  \mathbb{R}^{h\times w}$ to reduce the line of sight of the actual scene and remove some background noise.
The line of sight is limited of a fixed value generating a binary map $\textbf{\textit{X}}_{depthT}^{t} \in  \mathbb{N}^{h\times w}$, as follows:
\begin{equation}
\label{depth}
    {\textit{X}_{depthT_{\substack{i=0,...,h\\j=0,...,w}}}^{t}}(i,j)
    =\begin{cases} 0, & \mbox{if }(\textit{X}_{depth}^{t})_{i,j}\geq d_{depth} \\ 1, & \mbox{if }(\textit{X}_{depth}^{t})_{i,j}<d_{depth}
\end{cases}
\end{equation}
where $d_{depth}$ is a fixed experimental scalar.
Finally, exploiting an interception operation between the cumulative output $\hat{\textbf{\textit{X}}}_{cumSeg}^{t}$, computed in equation (\ref{sum_seg_maps}) and the binary map $\textbf{\textit{X}}_{depthT}^{t}$ previously generated, we obtain the pre-processed input  $\textbf{\textit{X}}_{ctrl}^{t}\in  \mathbb{R}^{h\times w}$ for the control algorithm:
\begin{equation}
    \textbf{\textit{X}}_{ctrl}^{t} = \hat{\textbf{\textit{X}}}_{cumSeg}^{t} \cap \textbf{\textit{X}}_{depthT}^{t}
\end{equation}
In the $\textbf{\textit{X}}_{ctrl}^{t}$ binary map $1$ stands for obstacles and $0$ free-space.
The segmentation-based control algorithm is developed building over the SPC algorithm presdented in \cite{aghi2021deep}. Indeed, we propose a simplified version of that control function to avoid useless conditional blocks and obtain a more real-time control algorithm.
\begin{algorithm}[t]
	\caption{Segmentation-based algorithm}
	\label{alg_seg_algorithm}
	\begin{algorithmic}[1]
	    \REQUIRE{\textbf{\emph{X}$_{ctrl}^{t}$}: Pre-processed segmented image}
	    \ENSURE{\textbf{$v_{x}$},\textbf{$\omega_{z}$}: Continuous control commands}
		\STATE {noise\_reduction\_function()}
		\FOR {i=0,$\cdots,w$}
	    	\STATE{ $ \textbf{\textit{c}}\leftarrow $sum\_colums($\textbf{\textit{X}}_{ctrl}^{t}$)}
		\ENDFOR
		\STATE{$ \textbf{\textit{zeros}}\leftarrow $ list\_zero\_clusters($\textbf{\textit{c}}$)}
		\STATE{$ \textbf{\textit{max\_cluster}}\leftarrow $ find\_max\_cluster($\textbf{\textit{zeros}}$)}
		\IF {cluster\_lenght($ \textbf{\textit{max\_cluster}}$) $\geq anomaly_{th}$}
		    \STATE{$v_{x}$,$\omega_{z} \leftarrow 0,0$} 
		\ELSE
		    \STATE{compute\_cluster\_center()}
		    \STATE{$v_{x}$,$\omega_{z} \leftarrow$ control\_function()} 
		\ENDIF
	\end{algorithmic}
\end{algorithm}
Algorithm \ref{alg_seg_algorithm} contains the pseudo-code of the proposed custom control, that starting from a pre-processed segmented image $\textbf{\textit{X}}_{ctrl}^{t}$, is responsible for computing the driving velocity commands. First, a noise reduction function gets rid of undesired noise in the bottom part of the cumulative segmentation mask $\textbf{\textit{X}}_{ctrl}^{t}$ due to grass on the terrain, that may be wrongly segmented by the neural network. To perform such operation, we compute the sum over rows of $\textbf{\textit{X}}_{ctrl}^{t}$ obtaining an array $\textbf{\textit{g}}_{noise}\in\mathbb{R}^{h}$, then we set ${\textbf{\textit{X}}_{ctrl}^{t}}_{(indices,:)}=0$, where $indices$ contains the matrix-row indices such that $\textbf{\textit{g}}_{noise}<{th}_{noise}$, with ${th}_{noise}=0.03 \cdot max(\textbf{\textit{g}}_{noise})$ as threshold. We perform such operation because, in an ideal segmentation mask there are no 1s at the top of the image and on the bottom, whilst the majority of them are supposed to be in the central belt.
After the noise reduction phase, we store the sum over columns of the obtained matrix $\textbf{\textit{X}}_{ctrl}^{t}$ in the array $\textbf{\textit{c}}\in  \mathbb{R}^{w}$, that contains the amount of segmented trees for each column. Therefore, every zero in $\textbf{\textit{c}}$ is a potential empty space where to route the mobile platform.
Then, we select the clusters of zeros in $\textbf{\textit{c}}$, which are the groups of consecutive zeros, in order to store them in the list $\textbf{\textit{zeros}}$.
Next, we look for the largest cluster of zeros $\textbf{\textit{max\_cluster}}$ and in case of the length of such cluster is over an empirically chosen threshold, $anomaly\_th=0.8\cdot w$, the driving commands are set to zero value, because it means the provided cumulative segmentation mask $\textbf{\textit{X}}_{ctrl}^{t}$ has more zeros than ones, that is an anomaly, so for safety reason the mobile platform is stopped. While, in case of no anomalies, we compute the cluster center that is given as input to the control function.
The identified cluster contains the obstacle-free space information that can be exploited to safely drive the mobile platform; as a consequence the linear and angular velocities are computed using the center of the selected cluster, which ideally corresponds to the center position of the row in front of the UGV. The desired velocities are obtained by means of two custom functions:
\begin{equation}
\label{eq1}
\omega_z =- \omega_{z,gain}  \cdot d
\end{equation}

\begin{equation}
\label{eq2}
v_{x} = v_{x,max} \cdot \left [ 1 - \left [\frac{d^{2}}{(\frac{w}{2})^{2}} \right ] \right ] 
\end{equation}
where $\omega_{z,gain}=0.01$ and $v_{x,max}=1.0$ are two constants which define the angular gain and maximum linear velocity of the mobile platform respectively, $w$ is the width of $\textbf{\textit{X}}_{ctrl}^{t}$ and \emph{d} is defined as:
    \begin{equation}
    \label{eq3}
    d = x_{c} - \frac{w}{2}
    \end{equation} 
with $x_{c}$ center coordinate of the selected cluster.
Equation (\ref{eq1}) represents the angular velocity control function, while equation (\ref{eq2}) has been used to compute the linear velocity, as in \cite{aghi2020local} and \cite{aghi2021deep}. 
Eventually, the control velocity commands sent to the actuators are smoothed using the Exponential Moving Average (EMA), formalized in equation (\ref{eq4}), in order to prevent the mobile platform from sharp motion.

\begin{equation}
    \label{eq4}
   \textbf{\textit{EMA}}_t=\textbf{\textit{EMA}}_{t-1} \cdot (1 - \alpha_{EMA}) + \begin{bmatrix} v_x \\ \omega_z  \end{bmatrix} \cdot \alpha_{EMA}
    \end{equation} 
where $t$ is the time step and $\alpha_{EMA}=0.18$ the multiplier for weighting the EMA, which value is found experimentally.

\subsection{Local Path Planning Policy}
Once the global path $\textbf{\emph{P}}$ and the start/end rows waypoints $\textbf{\emph{W}}_{ord}$ have been correctly generated, we exploit the provided information to locally navigate throughout the whole field. 

In case of lush vegetation and thick canopies that may distort GNSS signals inside the inter-row space, the local navigation problem is solved using the synergy of two different algorithms according to the kind of navigation requested in a determined place of the considered crop:
\begin{enumerate}
    \item Inside the inter-row space: we exploit the custom control algorithm, described in Section \ref{seg_based_section}, based on the segmentation information provided by the deep neural network, in order to overcome localization inaccuracies due to blocked GNSS signals by overgrown plant vegetation.
    \item Switch between different rows: we use the standard DWA, well described in \cite{dwa}, with fine-tuned parameters to safely switch between two rows following the circular path generated in Section \ref{sec:global_path_planning}, since outside the inter-row space a clear view of satellites and sky should be available.
\end{enumerate}
The choice of the suitable algorithm happens by comparing the estimated position by the EKF localization filter and the provided start/end rows waypoints; in case of start row recognition, the local path planning policy selects the segmentation-based control, otherwise it uses the DWA, as shown in Fig. \ref{fig:tot_system}. The comparison happens by computing a simple Euclidean distance between the start/end rows waypoint and the estimated positions. In case the calculated distance is lower than a threshold, $waypoint\_th=0.5$, the algorithm considers the waypoint as reached and selects the right local controller.

On the other hand, during specific year periods, when plant vegetation is not so dense and overgrown, the local path planning policy exploits only DWA to navigate throughout the whole field, thanks to a good view of both sky and satellites in every place of the crop, following the complete global path generated in Section \ref{sec:global_path_planning}.

\section{Experimental Results and Discussions}
\label{algorithm_testing}
In this section, we describe the main experiments conducted in simulation and real environments in order to better validate the algorithms. First, we discuss the synthetic datasets creation for the training of the two deep neural networks. Then, we illustrate the training process and the optimization techniques adopted to minimize inference costs. Finally, we conclude with the simulation and real environments evaluations.

 All the code have been developed ROS-compatible, in order to exploit some of the most used ROS packages in robotics research, as \textit{move\_base}\footnote{http://wiki.ros.org/move\_base} and \textit{robot\_localization}\footnote{http://docs.ros.org/en/melodic/api/robot\_localization/html/index.html} packages, that offer ready-to-use local planners and basic localization methods, respectively. Moreover, all tests have been performed using Ubuntu 18.04 and ROS Melodic.
\vspace{-7pt}
\subsection{Datasets creation}
As far as the two presented deep neural networks are concerned, they require to be trained on specific datasets according to the desired final application. Supervised learning training algorithms are the easiest to adopt with usually the best final results. However, they all require supervision by means of a labeled training dataset. That greatly affects costs and makes data collection complex and time-consuming. Therefore, we make exclusively use of synthetic data and domain randomization \cite{tobin2017domain} to enable affordable supervised learning and simultaneously bridge the domain gap between simulation and reality.

For DeepWay, a dataset of random occupancy grids is generated. All parameters such as number, orientation, depth, and length of the rows are randomly selected. The coordinates of starting and ending points of each row are generated by geometrical reasoning and the occupancy grid is obtained with circles of different radius for each location in between two corresponding points. Random holes in the rows are also created to increase the variability of the images. Ground truth waypoints are obtained to always lay inside the rows, since we experimentally found it helps the network prediction. Given two starting/ending points $\textbf{\textit{a}}$ and $\textbf{\textit{b}}$ of two successive rows, we compute the target waypoint as follows:

\begin{equation}
    \textbf{\textit{p}}_{\textbf{\textit{a}},\textbf{\textit{b}}} = \begin{bmatrix} 0 & \mp1 \\
                              \pm 1 & 0 \end{bmatrix}
              \frac{\textbf{\textit{a}}-\textbf{\textit{b}}}{2} + \frac{\textbf{\textit{a}}+\textbf{\textit{b}}}{2}
\end{equation}

that corresponds to a $\pm 90$ degrees rotation around the mean point, where the sign is selected to make the waypoint inside the row. We train DeepWay with a total of 3000 synthetic images and we validate it with 100 satellite images taken from the Google Maps database and manually annotated.
\begin{figure}[!ht]
\centering
\subfloat[]{\includegraphics[width=1.65in]{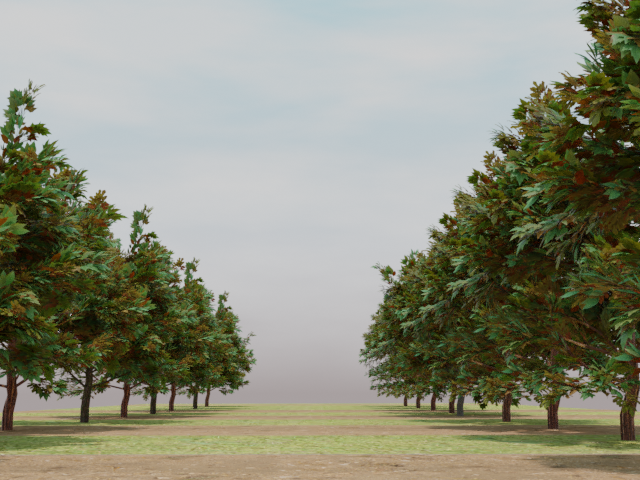}%
\label{fig:first_synt}}
\hfil
\subfloat[]{\includegraphics[width=1.65in]{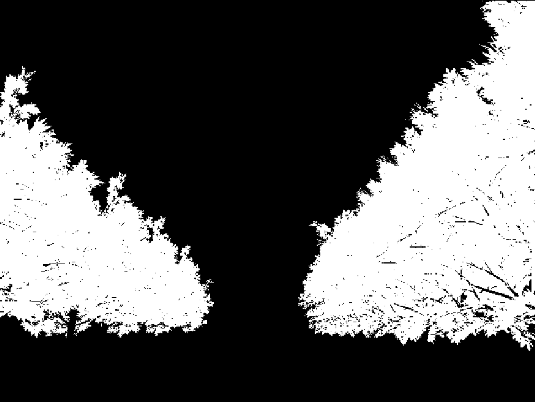}%
\label{fig:second_synt}}
\caption{An example of synthetic RGB image (a) and the corresponding segmentation mask (b).}
\label{fig:synthetic_dataset}
\end{figure}

On the other hand, the segmentation neural network requires a set of RGB images along with segmentation masks as inputs in order to be correctly trained. As a consequence, inspired by the work of Tejaswi Digumarti et al. \cite{semantic_seg_synthetic_images}, we generate synthetic RGB images coupled with the corresponding segmentation masks. For such purpose, we exploit Blender\footnote{https://www.blender.org/} 2.8, which is an open-source 3D computer graphics software compatible with Python language, and the Modular Tree\footnote{https://github.com/MaximeHerpin/modular\_tree/tree/blender\_28} addon to speed up the tree generation. We design four main scenarios: two single different trees, one group of heterogeneous trees, and one row-based scenario with various backgrounds and soils. Then, exploiting Python language compatibility of Blender, we write a script able to automatically capture RGB images and the corresponding segmentation masks of the scene from different positions with respect to the central reference frame and with different illuminations conditions in order to obtain as much as possible a random and complete synthetic dataset. An example of a synthetic RGB image, along with the corresponding segmentation mask, is shown in Fig. \ref{fig:synthetic_dataset}. Every single rendering takes about $30$ seconds, multiplied by a total of 2776 rendering, which is about $23$ hours of continuous work on a RTX 2080 GPU. That total number of images in conjunction with transfer learning allows to train a segmentation network with high generalization capabilities while minimizing generation data costs. The overall segmentation training dataset is composed of 2776 RGB synthetic images for training and 100 manually annotated images for testing, acquired in a real environment (Italy, Valle San Giorgio di Baone).
\vspace{-5pt}

\subsection{Networks training and optimization}
For training both networks, we employ the TensorFlow\footnote{https://www.tensorflow.org} 2
framework on a PC with 32-GB RAM, an Intel i7-9700K CPU, and an Nvidia 2080 Super GP-GPU.

DeepWay is trained following the methodology presented in \cite{mazzia2021deepway}, with an input dimension of $H=W=800$, and $N=2$ Residual Reduction modules, that result in a subsampling factor of $k=8$ and output dimensions of $U_H=U_W=100$. We select a kernel size of 5 and 16 filters for all the convolutional layers, except the first and last ones, that have kernel sizes of 7 and 3, respectively. These hyperparameters have been selected by performing a grid search over reasonable sets of values and adopting those that experimentally provided the best convergence. As loss function, a weighted mean squared function ($L_2$) is used to compensate the higher number of negative cells (i.e., with no waypoint in the target image) with respect to positive ones. We set these weights to 0.7 for the positive cells and 0.3 for the negative. The default distance threshold for the waypoints suppression algorithm is set to $d_{thr} = 8$ pixels, that is the minimum inter-row distance of our dataset, and the confidence threshold to $c_{thr} = 0.9$, in order to select the most confident predictions only. As in standard object detection algorithms, we adopt the Average Precision (AP) as metric for the prediction quality. We compute the AP at different distance ranges $d_r$. A prediction is considered a True Positive (TP) only if it falls within a distance $d_r$ form the target waypoint. On the 100 real-world images we reach an AP of 0.9794 with $d_r = 8 $ pixels, 0.9558 with 4 pixels and
0.7500 with 2 pixels. 

\begin{table}[!h]
\caption{Comparison between different devices' energy consumption and inference performances with graph optimization (G.O.) and weight precision (W.P.).}
\centering
\begin{tabular}{llllll}
\toprule
Device     & GO & WP   & Latency {[}ms{]} & E$_{net}$ {[}mJ{]} & Size {[}MB{]} \\ \hline
RTX 2080   & N  & FP32 & 28 $\pm$ 109        & 819             & 9.3          \\ 
           & Y  & FP32 & 0.1 $\pm$ 0.3    & 52              & 7.4          \\ 
           & Y  & FP16 & 0.1 $\pm$ 0.2      & 39              & 4.9           \\ 
Cortex-A57 & Y  & FP32 & 111 $\pm$ 0.9       & 166             & 4.2           \\
           & Y  & FP16 & 111 $\pm$ 2.3       & 165             & 2.2           \\ 
Cortex-A76 & Y  & FP32 & 55.4 $\pm$ 10.6     & 210             & 4.2           \\ 
           & Y  & FP16 & 65.3 $\pm$ 9.5      & 248             & 2.2           \\ \bottomrule
\end{tabular}
\label{tab:devices_comparison}
\end{table}
\vspace{-7pt}

Regarding the segmentation network, we train our model applying transfer learning to the selected backbone. Indeed, rather than using randomly initialized weights, we exploit MobileNetV3 variables derived from an initial training phase on the 1k classes and 1.3M
images of the ImageNet dataset \cite{deng2009imagenet}. Moreover, we pre-trained the overall segmentation network with Cityscapes \cite{cordts2016cityscapes}, a publicly available dataset with 30 different classes and 5000 fine annotated images. Finally, we adopt a strong data augmentation with random crops, brightness, saturation, contrast, rotation, and flips. All together, those techniques largely improve the final robustness of the model and its final generalization capability with a reduced number of training samples.
We train the network with stochastic gradient descent with a learning rate of 0.03 and Intersection over Unit (IoU) as loss function. The accuracy over the test set is 0.8 with a IoU of 0.46. In comparison, the accuracy of the validation set with 0.1 of the synthetic dataset is 0.86 with a domain gap of 0.08. Moreover, our experimentation shows that larger input sizes improve segmentation over the synthetic dataset, but greatly reduces accuracy over real images.

The trained network is optimized in order to reduce latency, inference cost, memory, and storage footprint. That is obtained with two distinct techniques: model
pruning and quantization. The first simplifies the topological structure, removing unnecessary parts of the architecture, and favors a more sparse model introducing
zeros to the parameter tensors. Subsequently, with
quantization, we reduce the precision of the numbers used
to represent model parameters from float32 to float16. That can be accomplished with a post-training quantization procedure.
In Table \ref{tab:devices_comparison} experimentation results with some reference architectures are summarized.

\begin{figure}[!ht]
\centering
\includegraphics[width=2.6in,height=1.4in]{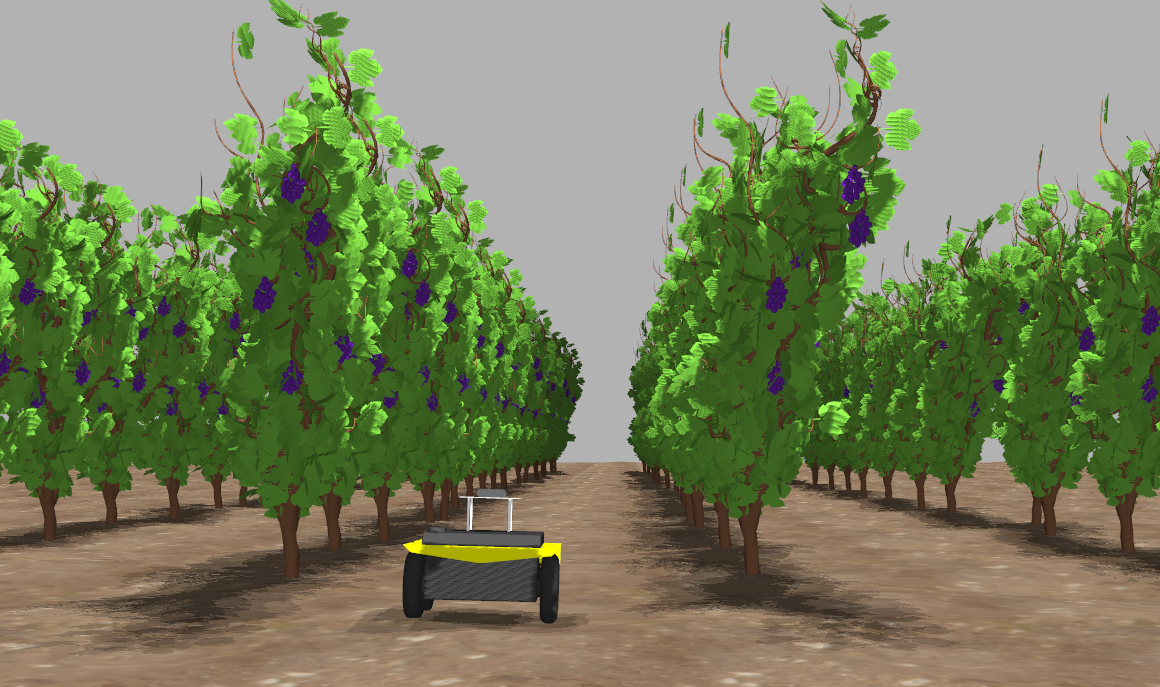}
\caption{A visual representation of the simulation environment.}
\label{fig:sim_vineyard}
\end{figure}

\subsection{Platform Hardware and Sensors Setup}
As mobile platform, we select the Jackal UGV by Clearpath Robotics, which can be briefly described as a small and weatherproof rover (IP62 code) with a 4x4 high-torque drivetrain. It is highly customizable and ROS-compatible allowing fast deployment and algorithm testing.
All the algorithms run on Jackal's onboard Mini-ITX PC with a CPU Intel Core i3-4330TE @2.4GHz and 4GB DDR3  RAM. For what concern the localization sensors, the RTK enabled GNSS receiver is the Piksi Multi by Swift Navigation\footnote{https://www.swiftnav.com/} mounted on an evaluation board, that provides easy input/output communication with the receiver (used acquisition rate: $10$ Hz), while the inertial measurements are provided by the MPU-9250 IMU, with an acquisition rate of about $100$ Hz. In addition, to get a front view of the environment, we select the Intel Realsense D455 RGBD camera, that provides frames at $30$ FPS and is mounted on the front part of Jackal's top plate. Finally, the odometry is provided by the on-board quadrature encoders that are able to run at $78000$ pulses/m.
As mentioned in Section \ref{system_overview}, the IMU and GNSS receiver data are fused by means of an EKF in order to obtain a global position estimate of the mobile platform time by time, described in terms of $x, y, yaw$. However, GNSS positioning is highly inaccurate, about $3m-5 m$, without implementing any corrections technique. As a consequence, we provide RTK corrections to Piksi Multi receiver, coming from the SPIN3 GNSS\footnote{https://www.spingnss.it/spiderweb/frmIndex.aspx} of Piemonte, Lombardia, and Valle d'Aosta, through the Internet. Then, the GNSS receiver directly uses such corrections to obtain more reliable and accurate global position estimates, with an error range of $[0.05,0.10]m$, in clear view of the sky and a good antenna position.
We exploit such corrections service because it is entirely free prior to an online subscription and it can send out RTK corrections through the Internet.
\vspace{-7pt}
\subsection{Simulation Environment Evaluation}
All the presented pipeline is tested in a simulation environment prior to real world tests in order to check the basic performances and perform a first experimental setting of various gains and thresholds. First, we build a custom simulation environment made of vine plants organized in rows, and a bumpy and uneven terrain, as shown in Fig. \ref{fig:sim_vineyard}, using the Gazebo\footnote{http://gazebosim.org/} simulator, that is ROS-compatible and open-source. Moreover, it provides advanced 3D graphics, dynamics simulation, and several plugins to simulate sensors, as GNSS, IMU, and cameras. 

Then, we compare the UGV trajectory obtained with the proposed methodology with a ground truth line, in order to evaluate different error metrics: Mean Absolute Error (MAE), Root Mean Square Error (RMSE) and Standard Deviation ($\sigma$), as shown in Table \ref{sim_test}. The ground truth is computed in two steps:
\begin{enumerate}
    \item manual annotation of GNSS simulated positions that correspond to an ideal global path: centered in the inter-row space and with a safe distance from crops switching between two different rows.
    \item linear interpolation of such points.
\end{enumerate}
We have performed six different tests in two different simulation environments (varying the vine plants positions and maintaining a row-based organization). The obtained performances are promising in terms of MAE, RMSE and $\sigma$, as shown in Table \ref{sim_test}. The worse attained results is stated by an RMSE=$0.265$ m and an MAE=$0.217$ m in the sixth test, while the best one is achieved in the first test with an RMSE=$0.089$ m and an MAE=$0.068$ m. Finally, the mean time to complete a test is about $4$ minutes with a maximum speed of $0.5$ m/s. All considered, the overall achieved performances are satisfactory taking into account the worse results are obtained in the last two tests where the vineyard rows are organized with a slight curvature shape.

\begin{table}[h]
\caption{Comparison of different error metrics in six different tests performed in two different simulation environments. The first row describes the number of visited rows in the corresponding test.}
\centering
\begin{tabular}{ccccccc}

     & T1  & T2 & T3 & T4 & T5 & T6 \\
\hline
N. rows    & $4$ & $4$ & $4$ & $4$ & $4$ & $4$ \\

Min. Error [m] & $0.001$ & $0.002$ & $0.002$ & $0.001$ & $0.002$ & $0.002$\\

Max. Error [m] & $0.726$ & $0.678$ & $0.600$ & $0.633$ & $1.21$ & $1.21$\\

MAE [m]    & $0.068$ & $0.085$ & $0.077$ & $0.082$ & $0.215$ & $0.217$\\

RMSE [m] & $0.089$ & $0.100$ & $0.092$ & $0.096$ & $0.263$ & $0.265$\\

$\sigma$ [m]   & $0.057$ & $0.053$ & $0.050$ & $0.050$ & $0.152$ & $0.152$ \\
\hline
           
\end{tabular}
\label{sim_test}
\end{table}

\subsection{Real Environment Evaluation}
\begin{figure}[!h]
\centering
\subfloat[Vineyard Row]{\includegraphics[width=1.65in,height=1.45in]{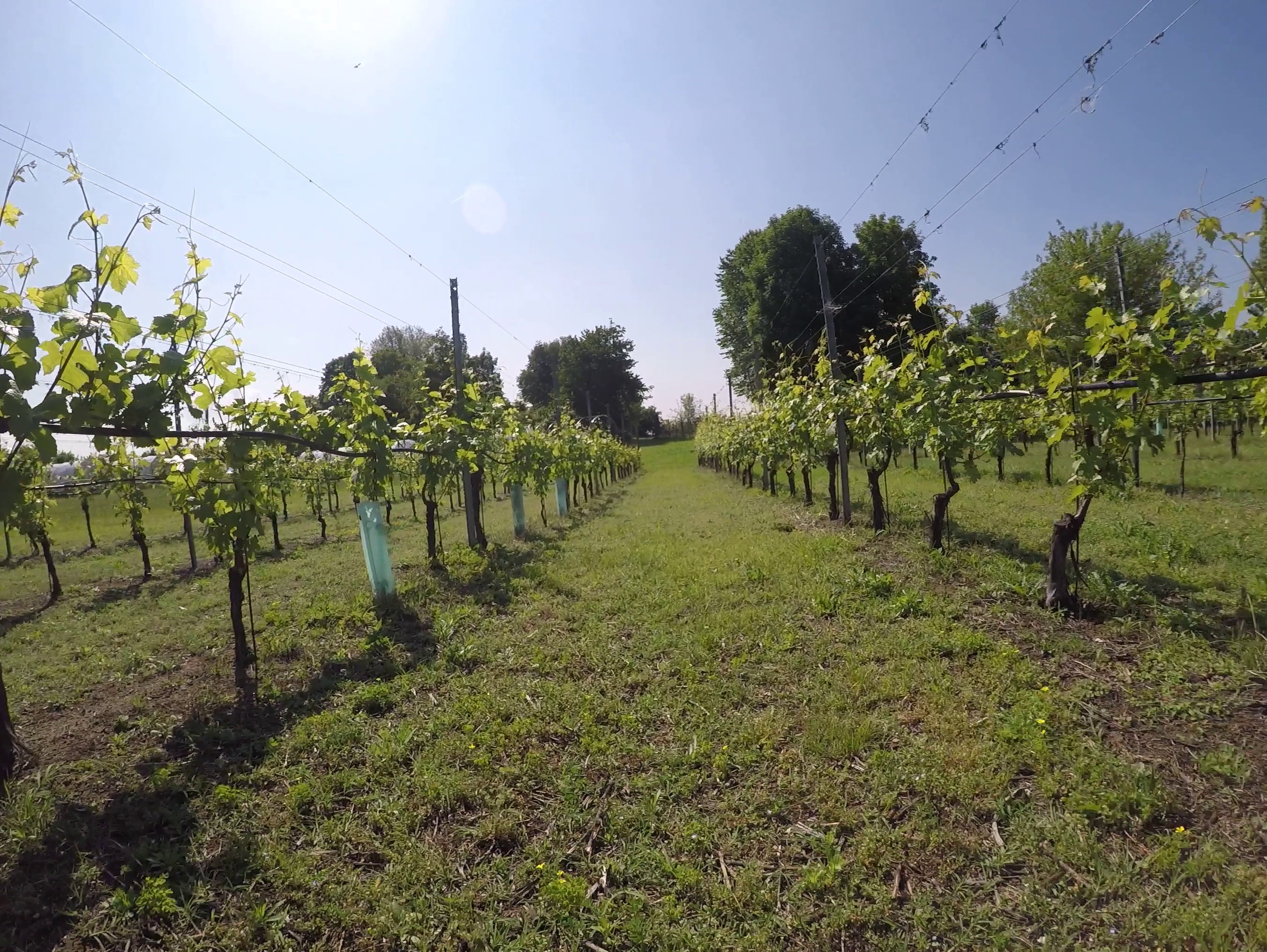}%
\label{fig:real_vineyard}}
\hfil
\subfloat[Pear Orchard Row]{\includegraphics[width=1.65in,height=1.45in]{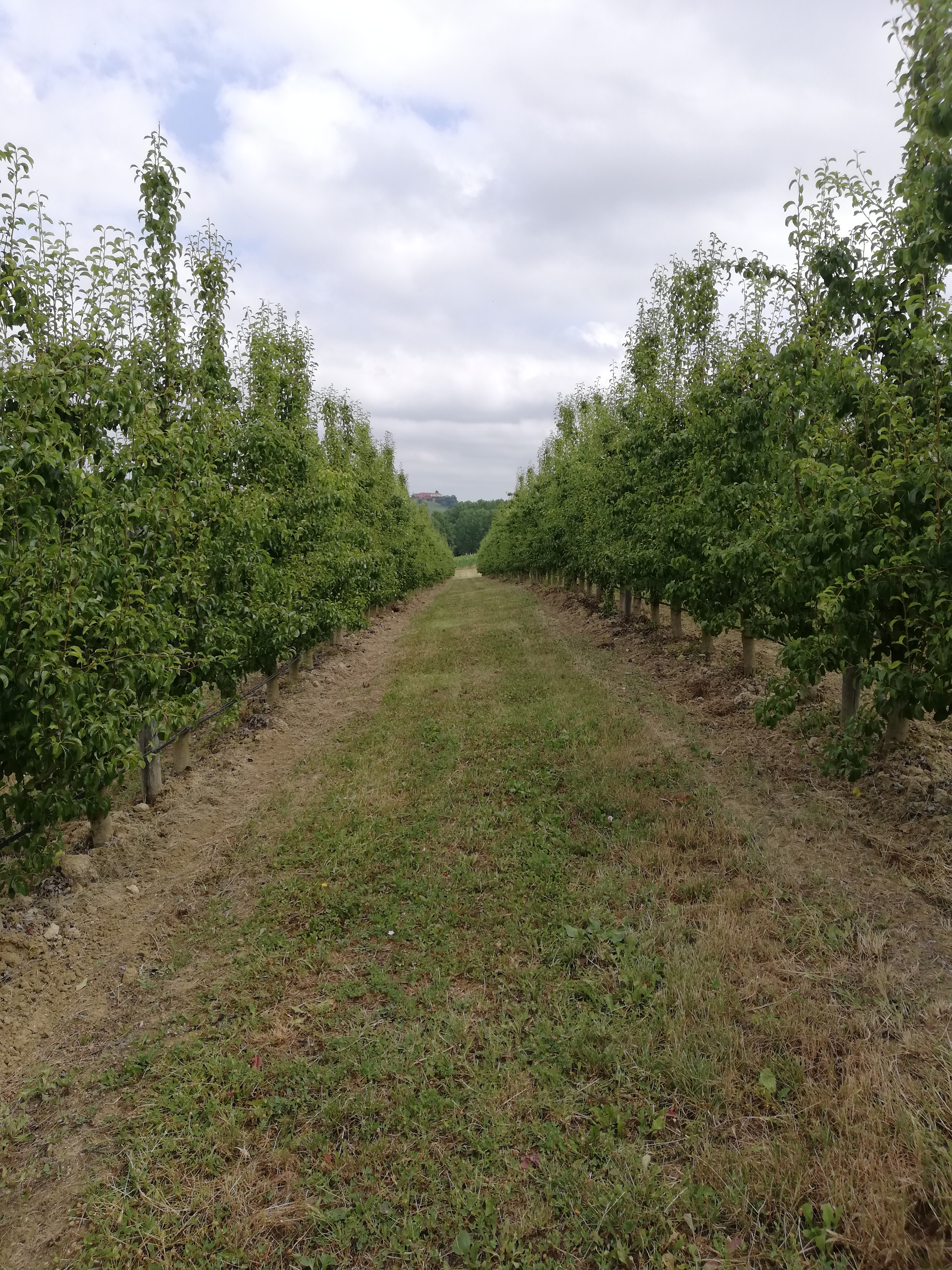}%
\label{fig:real_orchard}}
\caption{A visual representation of the real world testing environments.}
\label{fig:real_env_example}
\end{figure}

The overall system is extensively tested in two real environment scenarios with multiple experiments in different seasonal periods (Fig. \ref{fig:seasonal_variation}): a vineyard and a pear orchard, shown in Fig. \ref{fig:real_env_example}, for a total of more than 80 hours of experimentation from 9 a.m. to 6 p.m., but without particular adverse weather conditions (e.g., rain, snow, fog). Moreover, all the tests have been performed with the same hardware and software setup to obtain consistent data. The vineyard is located in Grugliasco and managed by the Department of Agricultural, Forestry and Food Sciences of Università degli studi di Torino (UNITO). In contrast, the pear orchard is located in Montegrosso d'Asti and managed by Mura Mura farm. The first scenario has an inter-row space of about $2.80m$ and a height of about $2.0 m$, while the second is organized in rows with an inter-row space of $4.50m$ and a height of about $3.0m$.

\begin{table}[h]
\caption{Comparison of different error metrics in three different tests performed in the pear orchard. The second column describes the number of visited rows in the corresponding test.}
\centering
\begin{tabular}{ccccccc}
Test     & \parbox[c]{0.8cm}{\centering N. rows} & \parbox[c]{0.8cm}{\centering Min. Error [m]}  & \parbox[c]{0.8cm}{\centering Max. Error [m]} & \parbox[c]{0.8cm} {\centering MAE [m]} & \parbox[c]{0.8cm} {\centering RMSE [m]} & \parbox[c]{0.75cm} {\centering $\sigma$ [m]}\\
\hline
\\
Test n. 1    & $4$ & $0.008$ & $1.395$ & $0.523$ & $0.627$ & $0.351$\\

Test n. 2    & $4$ & $0.002$ & $1.105$ & $0.457$ & $0.551$ & $0.315$\\

Test n. 3    & $2$ & $0.007$ & $1.320$ & $0.659$ & $0.755$ & $0.375$\\
\hline
           
\end{tabular}
\label{pear_orch_test}
\end{table}

\begin{table}[!h]
\caption{Comparison of different error metrics in three different tests performed in the vineyard. The second column describes the number of visited rows in the corresponding test.}
\centering
\begin{tabular}{ccccccc}

Test     & \parbox[c]{0.8cm}{\centering N. rows} & \parbox[c]{0.8cm}{\centering Min. Error [m]}  & \parbox[c]{0.8cm}{\centering Max. Error [m]} & \parbox[c]{0.8cm} {\centering MAE [m]} & \parbox[c]{0.8cm} {\centering RMSE [m]} & \parbox[c]{0.75cm} {\centering $\sigma$ [m]}\\
\hline
\\
Test n. 1    & $4$ & $0.013$ & $0.621$ & $0.296$ & $0.332$ &$0.160$\\

Test n. 2    & $6$ & $0.006$ & $0.598$ & $0.218$ & $0.240$ & $0.119$\\

Test n. 3    & $6$ & $0.003$ & $0.720$ & $0.204$ & $0.246$ & $0.145$\\
\hline
           
\end{tabular}
\label{vineyard_test}
\end{table}

The errors, described in Table \ref{pear_orch_test} and Table \ref{vineyard_test}, are computed comparing the RTK-GNSS positions provided by the Piksi Multi receiver and the global path provided to the navigation system. All of this is possible due to the high accuracy GNSS estimated positions, thanks to a clear view of the sky and a good position of the high-end antenna on the UGV. All the collected data are represented in latitude and longitude coordinates. However, for analysis purposes, they have been transformed in meters with respect to a known GNSS coordinate of the georeferenced occupancy grid map: the top left corner pixel.

\begin{figure}[!h]
\centering
\subfloat[Test n. 3 in the vineyard scenario]{\includegraphics[width=2.8in,height=2.4in]{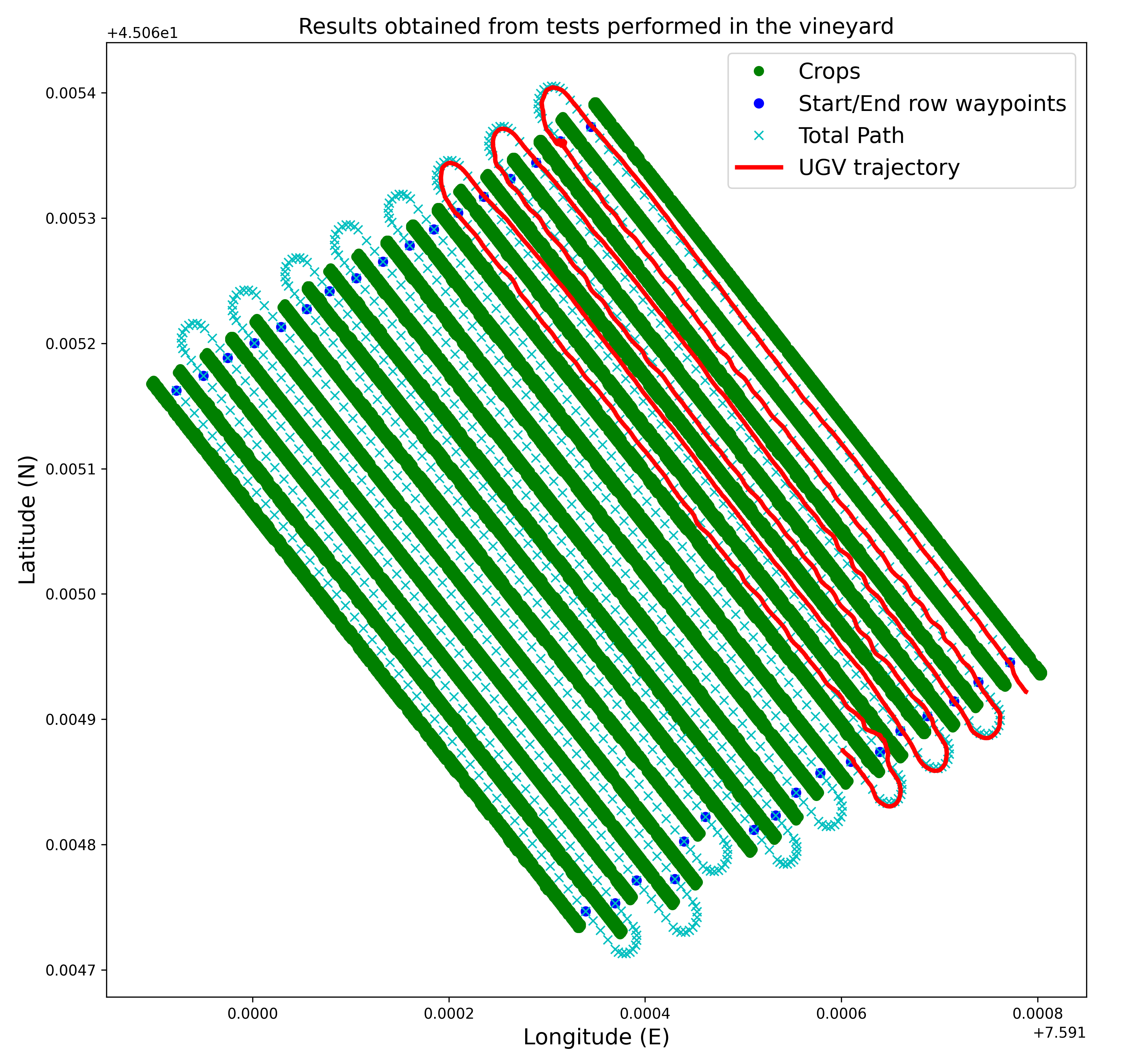}%
\label{fig:visual_res_vineyard}}
\hfil
\subfloat[Test n. 2 in the orchard scenario]{\includegraphics[width=2.8in,height=2.4in]{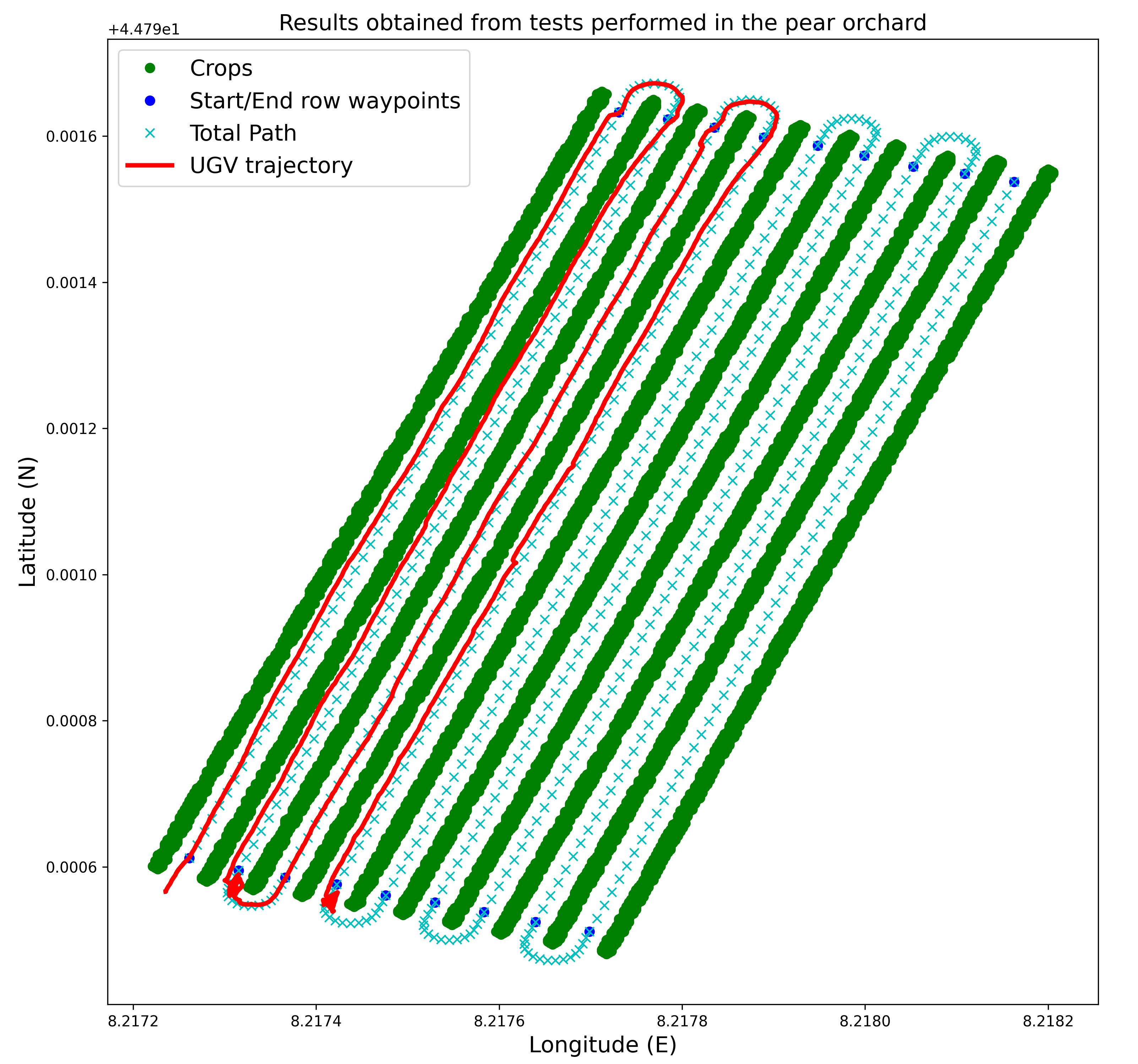}%
\label{fig:visual_res_orchard}}
\caption{A visual representation of the obtained results. Both images contain the path followed by the UGV (red line), the provided global path (cyan x), the start/end row waypoints (blue dots) and the crop (green dots).}
\label{fig:visual_results}
\end{figure}

Table \ref{pear_orch_test} and Table \ref{vineyard_test}, together with the visual representation of Fig. \ref{fig:visual_results}, shows some numerical results obtained by the proposed novel approach, and demonstrated that a methodology that exploits semantic segmentation along with a standard navigation approach based on the GNSS is able to provide complete and reliable navigation throughout the whole row-based crop. The results obtained in the pear orchard (Table \ref{pear_orch_test}) are slightly worse than the vineyard ones; however, this effect may be due to the greater inter-row space of the pear orchard with respect to the vineyard one. Eventually, the mean time of a single test is about $25$ minutes, while the maximum velocity of the UGV is $0.5$ m/s.  All considered, the overall achieved performances, in terms of MAE and RMSE, are adequate to the used low-cost sensors setup and demonstrate the effectiveness of our approach.

\section{Conclusion and Future Work}
\label{conclusions}
We presented a novel affordable algorithmic pipeline for autonomous navigation in row-based crops. Our methodology is a complete solution that covers all navigation stages, from global to local path planning, only relying on low-cost, low-range sensors. Moreover, the system efficiently tackles GNSS unreliability in presence of lush vegetation and thick canopies, allowing the platform to autonomously navigate in all seasonal periods. Finally, the adopted domain generalization and optimization techniques greatly make training and inference of deep neural network less time and computational costly. Further works will aim at assessing our proposed algorithmic pipeline onto a more cumbersome vehicle and introducing a collision avoidance algorithm in the segmentation-based control.


%



\section*{Acknowledgment}
This work has been developed with the contribution of the Politecnico di Torino Interdepartmental Centre for Service Robotics (PIC4SeR\footnote{\url{https://pic4ser.polito.it/}}) and SmartData@Polito\footnote{\url{https://smartdata.polito.it/}}.

\ifCLASSOPTIONcaptionsoff
  \newpage
\fi



\bibliographystyle{IEEEtran}
\bibliography{biblio}
%

%

\end{document}